\documentclass[a4paper,oneside,twocolumn]{article}
\usepackage[a4paper,left=2cm,right=1cm,top=2cm,bottom=2cm]{geometry}
\usepackage[usenames, dvipsnames]{xcolor}
\usepackage{booktabs}
\usepackage{graphicx}
\usepackage[breaklinks=true]{hyperref}
\usepackage{cleveref}
\usepackage{colortbl}
\usepackage{amssymb}	
\usepackage{balance}
\usepackage{comment}

\usepackage{tikz}
\usetikzlibrary{intersections}

\usepackage{listings}
\definecolor{dkgreen}{rgb}{0,0.6,0}
\definecolor{gray}{rgb}{0.5,0.5,0.5}
\definecolor{mauve}{rgb}{0.58,0,0.82}

\newcommand\code[1]{\lstinline[basicstyle=\ttfamily]|#1|}
\graphicspath{{images/}}
\usepackage{balance}

\begin{document}
\title{Reflection Invariance:\\an important consideration of image orientation}

\author{Craig Henderson, Ebroul Izquierdo\\
\small Multimedia and Vision Lab\\
\small Queen Mary University of London\\
\small \texttt{\{c.d.m.henderson, ebroul.izquierdo\}@qmul.ac.uk}}

\maketitle

\begin{abstract}
In this position paper, we consider the state of computer vision research with respect to invariance to the horizontal orientation of an image -- what we term \emph{reflection invariance}.
We describe why we consider reflection invariance to be an important property and provide evidence where the absence of this invariance produces surprising inconsistencies in state-of-the-art systems.
We demonstrate inconsistencies in methods of object detection and scene classification when they are presented with images and the horizontal mirror of those images.
Finally, we examine where some of the invariance is exhibited in feature detection and descriptors, and make a case for future consideration of reflection invariance as a measure of quality in computer vision algorithms.

\end{abstract}

\newpage
\section{Introduction}\label{sec:Introduction}


Human perception is invariant to horizontal reflection; they are equally able to recognise objects and scenes regardless of whether they are looking at an image as it has been taken as a photograph, or a horizontally reflected image, as if looking in a mirror.
We observe that computer vision algorithms are more sensitive to the reflection of an image and that invariance to this has not received any attention in contemporary research.
In this position paper, we introduce a property of \emph{reflection invariance}, specifically studying horizontal reflection as an introduction to the concept, although discussion is appropriate for general reflection about alternatives lines of symmetry.

We suggest \emph{reflection invariance} is an important property that should be considered in designing and implementing algorithms, and used as a metric in measuring the success of vision algorithms and applications.
Just as scale invariance seeks to neutralise the size of a feature to avoid bias in scale, we propose \emph{reflection invariance} to avoid bias in mirror reflection about an arbitrary axis. It is important that algorithms should be consistent in applications such as object recognition and scene classification, and we demonstrate that current state-of-the-art methods do not exhibit consistency when an image is reflected horizontally.

\section{Orientation and Reflection}\label{sec:SettingOut}
Low-level keypoint features describe a neighborhood of a few pixels, where the co-location of pixel intensities is an important attribute used to describe the feature. Most feature descriptors, including the most popular SIFT \cite{Lowe2004} and HoG \cite{Dalal2005}, use the orientation of pixel gradients in a color space or channel in some way to detect and represent distinct feature characteristics.
These algorithms are inherently sensitive to orientation, however others are sensitive only in practice, caused by poor implementation choices and mathematical rounding errors that accumulate to affect the result and cause dependence on image orientation.

A collection of descriptors can be composed to describe a distinctive pattern or region, such as in the popular \emph{Bag of Visual Words} method \cite{Sivic03}. In such a collection, the orientation of individual features \emph{relative to each other} is important, but the orientation of the collection as a whole is less significant.
As the scale of description increases further, orientation becomes less important and indeed becomes a limitation when considering high-level features in an image.
The significance of orientation can therefore be considered inversely proportional to the scale of description, with its influence diminishing with the increase in distance from the pixel detail (\Cref{{fig:pyramid}}).

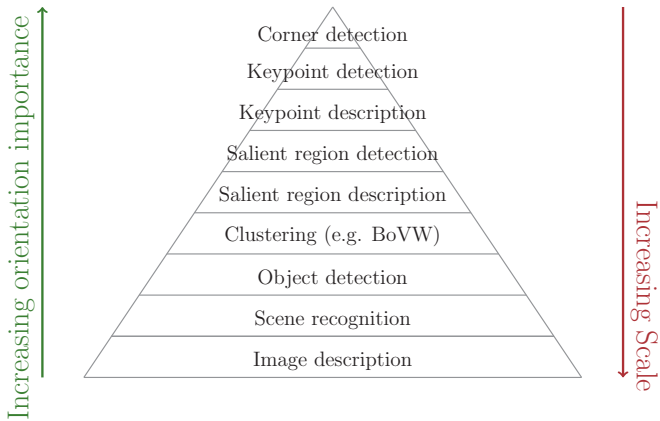
\begin{figure}
	\centering
	\resizebox{\linewidth}{!}{

%

\Large
\begin{tikzpicture}[thick,xscale=1, every node/.style={transform shape}]
\coordinate (A) at (-6,0) {};
\coordinate (B) at ( 6,0) {};
\coordinate (C) at (0,9) {};
\draw[name path=AC, Gray] (A) -- (C);
\draw[name path=BC, Gray] (B) -- (C);
\foreach \y/\A in {
	0/Image description,
	1/Scene recognition,
	2/Object detection,
	3/Clustering (e.g. BoVW),
	4/Salient region description,
	5/Salient region detection,
	6/Keypoint description,
	7/Keypoint detection,
	8/Corner detection,
	}
	{
    \path[name path=horiz] (A|-0,\y) -- (B|-0,\y);
    \draw[name intersections={of=AC and horiz,by=P},
          name intersections={of=BC and horiz,by=Q},
          Gray] (P) -- (Q)
    node[midway,above,Black] {\A};
}

\draw[ultra thick, OliveGreen, ->] (-7,0) -- (-7,9)
	node[anchor=south east, rotate=90, font=\fontsize{20}{22}\selectfont]
	{Increasing orientation importance};

\draw[ultra thick, Maroon, <-] (7,0) -- (7,9)
	node[anchor=south west, midway, rotate=-90, font=\fontsize{20}{22}\selectfont]
	{Increasing Scale};
\end{tikzpicture}}
	\caption{Pyramid of Scales and Orientation Significance: as the scale increases, the importance of orientation diminishes}
	\label{fig:pyramid}
\end{figure}

Reflection has the same scale of sensitivity as rota\-tional-orientation. Consider an example of scene recognition. A human would describe a city-scape scene, and identify a familiar city regardless of the horizontal reflection of the image; if the image is reflected about its vertical centre, this mirrored image would still be recognisable to a human and would not influence their description or identification. Computer vision algorithms, however, are more sensitive and often produce different results for these images.

\vspace{3mm}
The challenge is to generalize the description as the scale increases, with orientation becoming less relevant to the point where it is irrelevant at image scale.

\section{Reflection sensitivity in state-of-the-art methods}\label{sec:StateOfTheArt}
\subsection{Low level features}
Feature detectors fulfil the common need to identify interest points within an image. Information at these positions is extracted into a \emph{descriptor} -- a fixed length vector of numeric or binary values -- that can used, for example, to match similar features in applications such as image retrieval, alignment, stitching, and classification.

Many research papers combine the two stages of detection and description into a single step, but each are independent. The invariance properties of detectors and descriptors are important, and in work to date are consistent. An algorithm that provides for feature detection and feature description can provide invariance to scale, rotation, illumination or affine regions in both steps.

In considering invariance to horizontal reflection, we assess the two separately and propose that it is not necessary -- or even desirable -- for a method to be consistent in a reflection invariance in detection and description. The goal of feature detection is to find keypoints or regions in an image that contain \emph{interesting} information. The definition of \emph{interesting} is specific to the goal of the detector, but it is reasonable to expect that a location that is \emph{interesting} in an image should also be \emph{interesting} in the same image that is horizontally reflected.

\paragraph{Feature detectors}
To be reflection invariant, a feature detector must show that the set of keypoints or regions found in an image are equivalent to those found in the a mirror reflection of the image \cite{Henderson2015e}.
In that study, an analysis of feature detectors with respect to reflection invariance concluded that corner detectors are stable, and the most popular detectors SIFT and SURF are very unstable in detecting consistent feature points in images and their mirror reflections (\Cref{tab:Henderson2015}).

\begin{table}
	\centering
	\begin{tabular}{lc}
		\toprule
		Detector
		&	Invariant\\
		\midrule
		BRISK	&	No	\\
		FAST	&	Perfect	\\
		GFTT	&	Yes, after matching	\\
		HARRIS	&	Yes, after matching	\\
		ORB 	&	No	\\
		SIFT	&	No	\\
		STAR	&	Perfect	\\
		SURF	&	No	\\
		MSCR	&	Somewhat	\\
		MSER	&	Somewhat	\\
		\bottomrule
	\end{tabular}
	\vspace{1mm}
	\caption{Conclusions of the invariance characteristics of ten feature detectors from \cite{Henderson2015e}}
	\label{tab:Henderson2015}
\end{table}

\paragraph{Feature descriptors}
Conversely, the orientation of a feature is an important and discriminating attribute, and extracted descriptors should generally maintain local orientation so that established methods of feature matching, for example, can accurately measure the magnitude and position of a feature vector in high-dimensional space. However, reflection invariance in low-level descriptors can be especially useful for detecting intra-image lines of symmetry, such as water reflections in scene analysis. Research has explored reflection-invariant HoG \cite{Kanezaki2014} and, more frequently, SIFT-based methods such as RIFT \cite{Lazebnik2005}, MI-SIFT \cite{Ma2010} and MIFT \cite{Guo2012}. Generally, rotational invariance can be achieved by finding the dominant gradient and rotating the image patch so that the gradient is always in the same direction. RIFT, for example, divides normalized patches into four concentric rings of equal width, from each of which eight gradient orientation histograms are computed. The orientation is measured at each point relative to the direction pointing outward from the center, thus maintaining rotation invariance.

\subsection{Alignment and localization}
In a recent work, \cite{Yang2015} assessed object part localization and observed that the state-of-the-art methods augment the training set with mirrored images, but they did not result in bilaterally symmetric results. The authors introduced the term \emph{mirrorability} and a \emph{mirror error} that correlated with localization errors in human pose estimation and face alignment.

\subsection{Hough Forests}\label{sec:HoughForest}
Hough forests \cite{Gall2011} use a random forest framework \cite{Breiman2001} that are trained to learn a mapping from densely-sampled $D$-dimensional feature \emph{cuboids} to their corresponding votes in a Hough space $\mathcal{H} \subseteq \mathbb{R}^H$. The Hough space encodes the hypothesis $\mathbf{h}(c,\mathbf{x},s)$ for an object belonging to class $c\in C$ centred on $\mathbf{x}\in \mathbb{R}^D$ and with size $s$. The term \emph{cuboid} refers to a local image patch ($D = 2$) or video spatio-temporal neighborhood ($D = 3$) depending on the task.

Since their introduction in 2009 \cite{Gall2009}, Hough Forests have gained some interest in object detection tasks \cite{Barinova2012a,Ma2014b}.
Features are extracted from feature channels derived from an image, and are used to cast votes in Hough space. The standard set of 32 feature channels include \emph{Histograms of Oriented Gradients}-like features with $9$ bins using weighted orientations from a $5\times5$ neighborhood. The detected salient areas are therefore inherently sensitive to in-plane orientation and reflection.

\subsection{Deep learning}\label{sec:DeepLearning}

\begin{figure*}
	\centering
	\includegraphics[height=22mm]{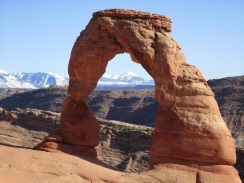}
	\includegraphics[height=22mm]{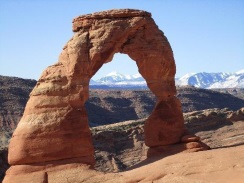}\hspace{1mm}
	\includegraphics[height=22mm]{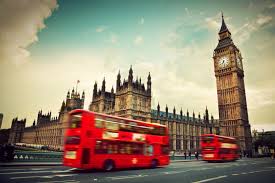}
	\includegraphics[height=22mm]{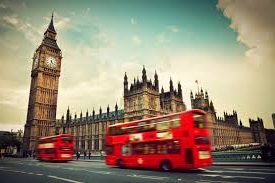}\hspace{1mm}\\
	\includegraphics[height=22mm]{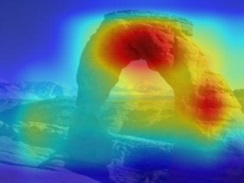}
	\includegraphics[height=22mm]{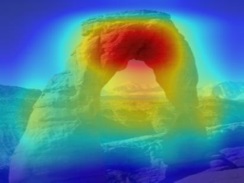}\hspace{1mm}
	\includegraphics[height=22mm]{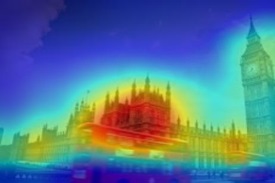}
	\includegraphics[height=22mm]{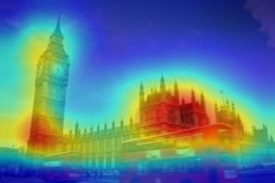}\\
	\vspace{3mm}
	\includegraphics[height=22mm]{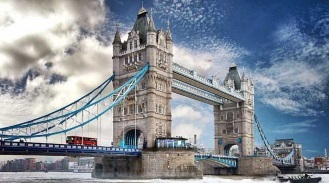}
	\includegraphics[height=22mm]{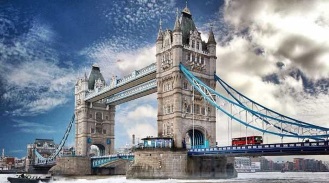}\hspace{1mm}
	\includegraphics[height=22mm]{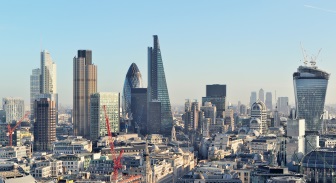}
	\includegraphics[height=22mm]{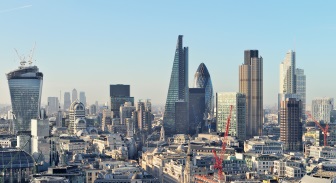}\hspace{1mm}\\
	\includegraphics[height=22mm]{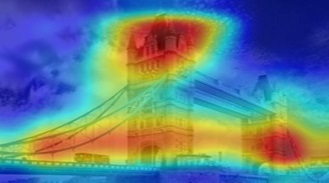}
	\includegraphics[height=22mm]{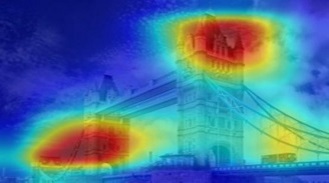}\hspace{1mm}
	\includegraphics[height=22mm]{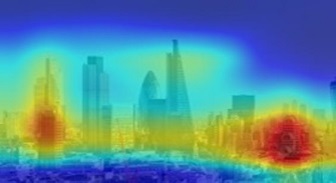}
	\includegraphics[height=22mm]{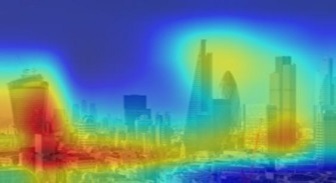}
	\caption{\emph{Informative regions} of images and their mirror, identified by \cite{Zhou2014a}. Note that the informative regions are not mirror images, suggesting the algorithms are sensitive to the horizontal orientation of the image.}
	\label{fig:MITInformativeRegions}
\end{figure*}

\begin{table*}
	\caption{Predictions from \cite{Zhou2014a} deep learning Scene Recognition system}
	\label{tab:MITInformativeRegions}
	\centering
	\begin{tabular}{
			>{\centering}m{27mm}
			>{\centering}m{16mm}
			>{\centering}m{30mm}
			>{\centering}m{65mm}
			c}
		\hline\noalign{\smallskip}
		&
		\textbf{Environment}&
		\textbf{Semantic categories}&
		\textbf{SUN scene attributes}&
		\textbf{Category}\\
		\noalign{\smallskip}\hline\noalign{\smallskip}
		\includegraphics[height=15mm]{4}&
		outdoor&
		rock\_arch:0.75, arch:0.24&
		naturallight, openarea, ruggedscene, climbing, rockstone, directsunsunny, dry, vacationingtouring, natural, warm&
		rock\_arch\\
		\includegraphics[height=15mm]{4r}&
		outdoor&
		rock\_arch:0.74, arch:0.25&
		naturallight, ruggedscene, rockstone, openarea, climbing, directsunsunny, dry, vacationingtouring, warm, natural&
		rock\_arch \\
		\noalign{\smallskip}\hline\noalign{\smallskip}
		\includegraphics[height=15mm]{westminster}&
		outdoor&
		tower:0.50, bridge:0.25, viaduct:0.12&
		man-made, clouds, openarea, naturallight, mostlyverticalcomponents, metal, vacationingtouring, nohorizon, directsunsunny, congregating&
		tower \\
		\includegraphics[height=15mm]{westminster_r}&
		outdoor&
		tower:0.50, bridge:0.25, viaduct:0.12&
		man-made, clouds, openarea, naturallight, mostlyverticalcomponents, metal, vacationingtouring, nohorizon, praying, directsunsunny&
		tower \\
		\noalign{\smallskip}\hline\noalign{\smallskip}
		\includegraphics[height=15mm]{20}&
		outdoor&
		skyscraper: 0.72, tower: 0.13, office\_building: 0.06&
		mostlyverticalcomponents, openarea, man-made, naturallight, directsunsunny, far-awayhorizon, clouds, metal, driving, transportingthingsorpeople&
		skyscraper \\
		\includegraphics[height=15mm]{20r}&
		outdoor&
		skyscraper:0.66, tower:0.13, office\_building:0.11&
		mostlyverticalcomponents, openarea, man-made, naturallight, directsunsunny, driving, transportingthingsorpeople, clouds, far-awayhorizon, metal&
		skyscraper \\
		\noalign{\smallskip}\hline\noalign{\smallskip}
		\includegraphics[height=15mm]{london}&
		outdoor&
		abbey:0.64, palace:0.16&
		man-made, clouds, openarea, mostlyverticalcomponents, naturallight, vacationingtouring, praying, nohorizon, electricindoorlighting, metal&
		abbey \\
		\includegraphics[height=15mm]{london_r}&
		outdoor&
		abbey:0.66, palace:0.15&
		clouds, man-made, openarea, mostlyverticalcomponents, naturallight, praying, vacationingtouring, nohorizon, metal, electricindoorlighting&
		abbey \\
		\noalign{\smallskip}\hline
	\end{tabular}
\end{table*}

While the recent adoption and development of neural network techniques have undoubtedly produced impressive results in computer vision tasks, and object and scene recognition in particular, they are not at all robust to variation in data.
Studies have shown that changing an image in a way imperceptible to humans can cause a deep neural network (DNN) to label the image as something else entirely \cite{Szegedy2014} and that it is easy to produce images that are completely unrecognizable to humans, but that state-of-the-art DNNs believe to be recognizable objects with 99.99\% confidence \cite{Nguyen2015}.

Recently, \cite{Zhou2014a} published research on a scene recognition system with an online demonstration\footnote{\url{http://places.csail.mit.edu/demo.html}}. \Cref{fig:MITInformativeRegions} shows a set of four images and their mirror reflections (top row) with the \emph{information regions} that the author's online demo produce. The information regions are salient areas that the system has identified in its quest to understand and describe an image. We find compelling the difference in the information regions and suggest that this demonstrates a bias to the horizontal orientation of the image.

\begin{table}
	\centering
	\caption{Object recognition results from the online \emph{Wolfram Language Image Identity Project}}
	\label{tab:WolframObjects}       
	\begin{tabular}{>{\centering\bfseries}m{16mm}
			>{\centering}m{26mm}
			>{\centering\arraybackslash}m{26mm}}
		\hline\noalign{\smallskip}
		\textbf{Resolution}& \textbf{Original}& \textbf{Mirror}\\
		\hline\noalign{\smallskip}
		$550 \times 412$	& \includegraphics[height=15mm]{4} arch	& \includegraphics[height=15mm]{4r} arch\\
		$244 \times 183$	& \includegraphics[height=15mm]{4} broken arch	& \includegraphics[height=15mm]{4r} arch\\
		\hline\noalign{\smallskip}
		$736 \times 490$	& \includegraphics[height=15mm]{westminster} fire truck	& \includegraphics[height=15mm]{westminster_r} fire truck\\
		$275 \times 183$	& \includegraphics[height=15mm]{westminster} building	& \includegraphics[height=15mm]{westminster_r} building\\
		\hline\noalign{\smallskip}
		$607 \times 338$	& \includegraphics[height=15mm]{20} bascule	& \includegraphics[height=15mm]{20r} church\\
		$329 \times 183$	& \includegraphics[height=15mm]{20} church	& \includegraphics[height=15mm]{20r} church\\
		\hline\noalign{\smallskip}
		$4370 \times 2383$	& \includegraphics[height=15mm]{london} oil refinery	& \includegraphics[height=15mm]{london_r} industrial park\\
		$336 \times 183$	& \includegraphics[height=15mm]{london} oil refinery	& \includegraphics[height=15mm]{london_r} oil refinery\\
		\noalign{\smallskip}\hline
	\end{tabular}
\end{table}

\Cref{tab:MITInformativeRegions} shows the detailed results of the scene recognition. The system determines the environment, semantic categories and SUN scene attributes \cite{Xiao2014}. The category column summarizes the highest scoring semantic category. Despite the differences in salient areas of the images, the overall categorization has not been affected. Each image and its mirror image are categorized the same in these examples. However, there are differences in the detail, which illustrate inconsistencies that, in boundary cases, could change the categorization. The semantic categories are rated with a likelihood.
The Rock Arch -- a stock image from the author's own demonstration -- reduces in likelihood by 0.01 in the mirror image, the Palace of Westminster\footnote{\url{https://s-media-cache-ak0.pinimg.com/736x/d2/5a/0e/d25a0ed9bb2e788ae9c9ec59cc52670c.jpg}} is classified exactly the same in each pair, Tower Bridge -- another stock image from the author's own demonstration -- appears less like a skyscraper and more like an office building in the reflected image than in the original, and the City of London skyline\footnote{\url{http://upload.wikimedia.org/wikipedia/commons/d/da/The_City_London.jpg}} increases its likelihood of being an abbey in the reflection image. The inconsistency in the ratings, albeit small, further strengthen our resolve that computer vision systems are commonly bias to image horizontal orientation.
It is also interesting to note that images from the author's own demonstration score higher in the semantic categorization than images from other sources.

We used a second neural network based object recognition system, \emph{The Wolfram Language Image Identification Project}\footnote{\url{https://www.imageidentify.com/}}, to test classification of our images, this time using different sizes of the same image. \Cref{tab:WolframObjects} shows the results; the Rock Arch is classified differently in its original orientation at a small scale, the Palace of Westminster was classified consistently at each scale, Tower Bridge is classified differently in its original orientation at a large scale and the London Skyline is classified differently in its mirror orientation at a large scale. These results show that this system is sensitive to scale, and that the scale change also influences the invariance to horizontal reflection.

\begin{figure}
	\centering
	\includegraphics[width=41mm]{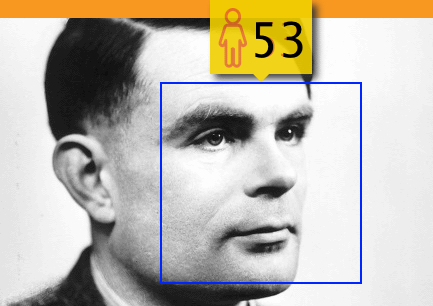}\hspace{1mm}
	\includegraphics[width=41mm]{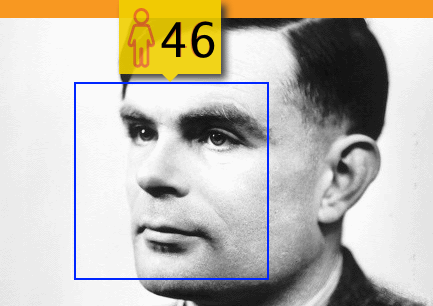}\\\vspace{2mm}
	\includegraphics[width=41mm]{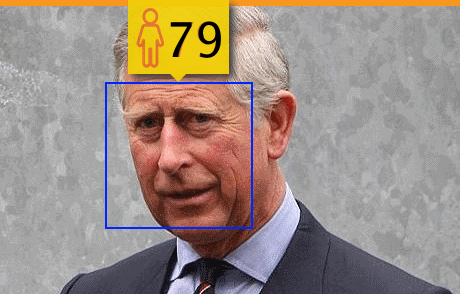}\hspace{1mm}
	\includegraphics[width=41mm]{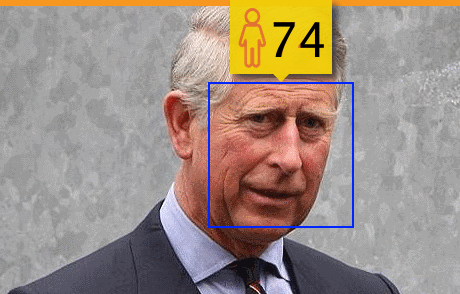}
	\caption{Microsoft's How-Old.Net demonstration attempts to guess a person's age from a photograph image. These two examples demonstrate that the system is sensitive to image orientation -- and not head orientation -- as the ages are quite different for each pair.}
	\label{fig:MicrosoftHowOld}
\end{figure}

Finally, Microsoft's much publicised How-Old.net\footnote{\url{http://how-old.net}} asks ``\emph{How Old Do I Look?}'' and uses machine learning to guess the answer to the question from a photograph. We used photographs of Alan Turing\footnote{\url{https://kpfa.org/wp-content/uploads/2015/05/Dr-Alan-Turing-2956483.jpg}} and Prince Charles\footnote{\url{http://i.telegraph.co.uk/multimedia/archive/01422/princeCharles_1422434c.jpg}} and observed the difference in age that was guessed for each image and its reflection (\Cref{fig:MicrosoftHowOld}). In both cases, the ages decreased in the reflected image (\emph{right}), despite the orientation of the head being different in each case.

This inconsistency in results is perhaps more surprising as the image orientation affects the guess of the person's age, but the system does not appear to be intrinsically biased towards the orientation of the head itself. On close examination, the bounding boxes of the identified \emph{faces} are different sizes -- smaller in the reflected image in both cases -- by 5 pixels in each \emph{x}- and \emph{y}-axis in the case of the photograph of Alan Turing and 1 pixel in each axis in the case of Prince Charles. The detected face of Alan Turing is in a consistent corner position relative to the visible ear, and the detected face of Prince Charles is consistent in the opposite top corner. We therefore conclude that the face detection algorithm used in the system is sensitive to head orientation and this may affect the subsequent learned system of age estimation, which may or may not be orientation-sensitive itself.

\section{Algorithms and Implementations}\label{sec:Algorithms}
Many algorithms described in the research literature -- especially saliency based feature detectors -- are not inherently sensitive to orientation. Nonetheless, no mention is made of reflection invariance in the papers, suggesting a general unawareness of this property.
Consequently, we have observed several cases where commonly used, freely available code -- including reference implementations from the original authors -- have an invariance worsened by, or \emph{caused by}, choices made in the implementation. For example, algorithms that use a Difference-of-Gaussian pyramid for sub-pixel feature detection can inadvertently increase their reflection dependence by the use of 32-bit floating point arithmetic for intermediate calculations.
Using the popular OpenCV \cite{opencv_library} tool kit for C++, we tested the\\ \code{GaussianBlur()} function that convolves an image with a specified Gaussian kernel. We found that using 32-bit arithmetic produces reflection-sensitive convolutions for many images that we tested (not shown), but using 64-bit arithmetic all convolutions of our test images were reflection invariant (\Cref{fig:OpenCVGaussianBlur}).

Conceptually, one would expect salient regions to be less biased to horizontal orientation, because they use neighborhood color and intensity measures and are less dependent on pixel gradients. However, common implementations of salient region detectors such as \emph{maximally stable extremal regions} (MSER) \cite{Forssen2007a} can suffer in the initial step of the algorithm blurring the image with a Gaussian kernel. In their saliency detector reference implementation\footnote{\url{https://github.com/MingMingCheng/CmCode}}, \cite{Cheng2011a} exhibit orientation sensitivity due to many reasons including floating point errors in color quantization which are realized differently dependent on the order in which the data is processed, which is determined by the image orientation. Increasing floating point arithmetic to double-precision 64-bit calculations correct the quantization sensitivity to reflection invariance.

\begin{figure}
	\lstset{numbers=left,xleftmargin=1em}
	\begin{lstlisting}
	using namespace cv;
	
	Mat src = imread("image.png", CV_LOAD_IMAGE_GRAYSCALE);
	
	Mat fpt;
	src.convertTo(fpt, CV_32F, SIFT_FIXPT_SCALE, 0);
	
	Mat fpt_r;
	flip(fpt, fpt_r, 1);
	
	auto sigma = 1.24899971;
	GaussianBlur(fpt, fpt, Size(), sigma, sigma);
	GaussianBlur(fpt_r, fpt_r, Size(), sigma,sigma);
	
	assert(countNonZero(fpt - fpt_r) == 0);
	\end{lstlisting}
	\caption{Example \code{C++} code to test reflection invariance of a Gaussian filter in OpenCV. Using 32-bit floating point arithmetic -- \code{CV\_32F} on line 6 -- will often result in an assertion failure on line 15 indicating that a Gaussian filter on a horizontally flipped image does not produce the same as the result as applying the same filter to the original image. Changing to use 64-bit double precision arithmetic -- \code{CV\_64F} -- produces identical results on all of our test images, with no assertion failures.}
	\label{fig:OpenCVGaussianBlur}
\end{figure}

\section{Conclusion}\label{sec:Conclusion}
We have proposed \emph{reflection invariance} to be an important consideration when designing and implementing algorithms.
In citing contemporary research projects, we have demonstrated inconsistencies in applications of scene classification, object detection and age-guessing when presented with images and their horizontal reflections. We have described where some of the sensitivity is exhibited in feature detection and descriptors and applications of alignment and localization. 
We therefore urge researchers to consider reflection invariance when designing and implementing  algorithms, and suggest reflection consistency should be introduced as a measurement of success of algorithms and their improvement over state-of-the-art.

\section*{Acknowledgements}
This work is funded by the European Union's Seventh Framework Programme, specific topic ``framework and tools for (semi-)automated exploitation of massive amounts of digital data for forensic purposes'', under grant agreement number 607480 (LASIE IP project). The authors also extend their thanks to the Metropolitan Police at Scotland Yard, London, UK, for the supply of and permission to use CCTV images.

\balance
\bibliographystyle{ieeetr}

\begin{thebibliography}{10}
	
	\bibitem{Lowe2004}
	D.~G. Lowe, ``{Distinctive image features from scale-invariant keypoints},''
	{\em International Journal of Computer Vision}, vol.~60, pp.~91--110, Nov.
	2004.
	
	\bibitem{Dalal2005}
	N.~Dalal and B.~Triggs, ``{Histograms of Oriented Gradients for Human
		Detection},'' in {\em 2005 IEEE Computer Society Conference on Computer
		Vision and Pattern Recognition (CVPR'05)}, vol.~1, pp.~886--893, IEEE, 2005.
	
	\bibitem{Sivic03}
	J.~Sivic and A.~Zisserman, ``{Video Google: a text retrieval approach to object
		matching in videos},'' in {\em Proceedings Ninth IEEE International
		Conference on Computer Vision}, vol.~2, pp.~1470--1477, Oct. 2003.
	
	\bibitem{Henderson2015e}
	C.~Henderson and E.~Izquierdo, ``{Low level feature detectors and their
		invariance to Bilateral Symmetry}.'' 2015.
	
	\bibitem{Kanezaki2014}
	A.~Kanezaki, Y.~Mukuta, and T.~Harada, ``{Mirror reflection invariant HOG
		descriptors for object detection},'' in {\em 2014 IEEE International
		Conference on Image Processing (ICIP)}, pp.~1594--1598, IEEE, Oct. 2014.
	
	\bibitem{Lazebnik2005}
	S.~Lazebnik, C.~Schmid, and J.~Ponce, ``{A sparse texture representation using
		local affine regions},'' Aug. 2005.
	
	\bibitem{Ma2010}
	R.~Ma, J.~Chen, and Z.~Su, ``{MI-SIFT},'' in {\em Proceedings of the ACM
		International Conference on Image and Video Retrieval - CIVR '10}, (New York,
	New York, USA), p.~228, ACM Press, July 2010.
	
	\bibitem{Guo2012}
	X.~Guo and X.~Cao, ``{MIFT: A framework for feature descriptors to be mirror
		reflection invariant},'' {\em Image and Vision Computing}, vol.~30, no.~8,
	pp.~546--556, 2012.
	
	\bibitem{Yang2015}
	H.~Yang and I.~Patras, ``{Mirror, mirror on the wall, tell me, is the error
		small?},'' in {\em CVPR 2015}, 2015.
	
	\bibitem{Gall2011}
	J.~Gall, A.~Yao, N.~Razavi, L.~{Van Gool}, and V.~Lempitsky, ``{Hough forests
		for object detection, tracking, and action recognition},'' {\em IEEE
		Transactions on Pattern Analysis and Machine Intelligence}, vol.~33,
	pp.~2188--2202, 2011.
	
	\bibitem{Breiman2001}
	L.~Breiman, ``{Random forests},'' {\em Machine Learning}, vol.~45, pp.~5--32,
	2001.
	
	\bibitem{Gall2009}
	J.~Gall and V.~Lempitsky, ``{Class-specific hough forests for object
		detection},'' in {\em 2009 IEEE Computer Society Conference on Computer
		Vision and Pattern Recognition Workshops, CVPR Workshops 2009},
	pp.~1022--1029, 2009.
	
	\bibitem{Barinova2012a}
	O.~Barinova, V.~Lempitsky, and P.~Kholi, ``{On detection of multiple object
		instances using hough transforms},'' {\em IEEE Transactions on Pattern
		Analysis and Machine Intelligence}, vol.~34, pp.~1773--84, Sept. 2012.
	
	\bibitem{Ma2014b}
	K.~Ma and J.~Ben-Arie, ``{Compound Exemplar Based Object Detection by
		Incremental Random Forest},'' in {\em 2014 22nd International Conference on
		Pattern Recognition}, pp.~2407--2412, IEEE, Aug. 2014.
	
	\bibitem{Zhou2014a}
	B.~Zhou, A.~Lapedriza, J.~Xiao, A.~Torralba, and A.~Oliva, ``{Learning Deep
		Features for Scene Recognition using Places Database},'' in {\em Advances in
		Neural Information Processing Systems 27} (Z.~Ghahramani, M.~Welling,
	C.~Cortes, N.~D. Lawrence, and K.~Q. Weinberger, eds.), pp.~487--495, Curran
	Associates, Inc., 2014.
	
	\bibitem{Szegedy2014}
	C.~Szegedy, W.~Zaremba, I.~Sutskever, J.~Bruna, D.~Erhan, I.~Goodfellow, and
	R.~Fergus, ``{Intriguing properties of neural networks}.'' Dec. 2014.
	
	\bibitem{Nguyen2015}
	A.~Nguyen, J.~Yosinski, and J.~Clune, ``{Deep Neural Networks are Easily
		Fooled: High Confidence Predictions for Unrecognizable Images},'' in {\em
		CVPR 2015}, 2015.
	
	\bibitem{Xiao2014}
	J.~Xiao, K.~a. Ehinger, J.~Hays, A.~Torralba, and A.~Oliva, ``{SUN Database:
		Exploring a Large Collection of Scene Categories},'' {\em International
		Journal of Computer Vision}, 2014.
	
	\bibitem{opencv_library}
	G.~R. Bradski, ``{The OpenCV Library},'' {\em Dr. Dobb's Journal of Software
		Tools}, 2000.
	
	\bibitem{Forssen2007a}
	P.-E. Forss\'{e}n and D.~G. Lowe, ``{Shape descriptors for maximally stable
		extremal regions},'' in {\em Proceedings of the IEEE International Conference
		on Computer Vision}, pp.~1--8, 2007.
	
	\bibitem{Cheng2011a}
	M.-M. Cheng, G.-X. Zhang, N.~J. Mitra, X.~Huang, and S.-M. Hu, ``{Global
		contrast based salient region detection},'' in {\em CVPR 2011}, pp.~409--416,
	IEEE, June 2011.
	
\end{thebibliography}

\end{document}